\documentclass[10pt,journal,compsoc]{IEEEtran}

\usepackage{xcolor,soul,framed} %,caption

\colorlet{shadecolor}{yellow}
\usepackage[pdftex]{graphicx}
\graphicspath{{../pdf/}{../jpeg/}}
\DeclareGraphicsExtensions{.pdf,.jpeg,.png}

\usepackage[cmex10]{amsmath}
\usepackage{array}
\usepackage{mdwmath}
\usepackage{mdwtab}
\usepackage{eqparbox}
\usepackage{url}
\usepackage{graphicx}
\usepackage{amsmath}
\usepackage{amsfonts}
\usepackage{algorithm}
\usepackage{algpseudocode}
\usepackage{multirow}
\usepackage{footnote}
\usepackage{bm}
\usepackage{makecell}
\usepackage{pdflscape}
\usepackage{afterpage}
\usepackage{bbding}
\usepackage{lscape}
\usepackage{booktabs}
\usepackage{tikz}

\usepackage{pgfplots}
\pgfplotsset{compat=1.12}
\usepackage{subcaption}
    \pgfplotsset{
        layers/my layer set/.define layer set={
            background,
            main,
            foreground
        }{ },
        set layers=my layer set,
    }

\usepackage[square,sort,comma,numbers]{natbib} % multiple citation or 
\usepackage[top=0.4in,bottom=0.4in,left=0.5in,textwidth=7.6in]{geometry}

\usepackage[pagebackref=false,breaklinks=false,linkcolor=black,anchorcolor=black, citecolor=black,colorlinks,bookmarks=true]{hyperref}
\usepackage{amsfonts}

\usepackage{pifont}

\begin{document}
% \title{Exploring Sparsity for Fast Inference of Unsupervised Moving Vehicle Detection in Satellite Videos}

\title{Highly Efficient and Unsupervised Framework for Moving Object Detection in Satellite Videos}

\author{Chao~Xiao, Wei~An, Yifan~Zhang, Zhuo~Su,  Miao~Li, Weidong~Sheng, Matti ~Pietik{\"a}inen, Li~Liu% <-this % stops a space  Zhijie~Chen,
\IEEEcompsocitemizethanks{
\IEEEcompsocthanksitem This work was supported by the National Key Research and Development Program of China No. 2021YFB3100800, the National Natural Science Foundation of China under Grant 62376283, and  the Key Stone grant (JS2023-03) of the National University of Defense Technology (NUDT). C. Xiao (xiaochao12@nudt.edu.cn), W. An, Y. Zhang, M. Li, W. Sheng (shengweidong1111@sohu.com), and L. Liu (liuli\_nudt@nudt.edu.cn) are with the College of Electronic Science and Technology, National University of Defense Technology, Changsha 410073, China. Z. Su and M. Pietik{\"a}inen are with the Center for Machine Vision and Signal Analysis (CMVS) at the University of Oulu, Finland. 
% Z.~Chen is with the National Airspace Technology Key Laboratory, Beijing 100085, China.
Li Liu and Weidong Sheng are the corresponding authors.}}

\markboth{Submitted to IEEE Transactions on Pattern Analysis and Machine Intelligence}%
{Xiao \MakeLowercase{\textit{et al.}}: Moving object detection}

\IEEEtitleabstractindextext{%
\begin{abstract}
Moving object detection in satellite videos (SVMOD) is a challenging task due to the extremely dim and small target characteristics. Current learning-based methods extract spatio-temporal information from multi-frame dense representation with labor-intensive manual labels to tackle SVMOD, which needs high annotation costs and contains tremendous computational redundancy due to the severe imbalance between foreground and background regions. In this paper, we propose a highly efficient unsupervised framework for SVMOD. Specifically, we propose a generic unsupervised framework for SVMOD, in which pseudo labels generated by a traditional method can evolve with the training process to promote detection performance. Furthermore, we propose a highly efficient and effective sparse convolutional anchor-free detection network by sampling the dense multi-frame image form into a sparse spatio-temporal point cloud representation and skipping the redundant computation on background regions. Coping these two designs, we can achieve both high efficiency (label and computation efficiency) and effectiveness. Extensive experiments demonstrate that our method can not only process 98.8 frames per second on $1024 \times 1024$ images  but also achieve state-of-the-art performance. The relabeled dataset and code are available at \url{https://github.com/ChaoXiao12/Moving-object-detection-in-satellite-videos-HiEUM}.
\end{abstract}

}

\maketitle

\IEEEdisplaynontitleabstractindextext

\IEEEpeerreviewmaketitle

\IEEEraisesectionheading{\section{Introduction}\label{sec:introduction}}

As satellite video technology has progressed over the last decade, video satellites capable of observing the Earth continuously with high temporal resolution have emerged as an essential tool for Earth observation.  Consequently, more and more video satellites and satellite constellations will be launched. Satellite videos can be accessed more easily than before, enabling more research opportunities and applications. Therefore, satellite video intelligent interpretation has become a pressing need, and the new frontier in remote sensing \cite{li2023recent}. Moving object detection in satellite videos (SVMOD), aiming to locate objects of interest, is a fundamental task in satellite video interpretation and has various applications including military surveillance, transportation planning, and public security \cite{ao2019needles,zhang2019error, xiao2023incorporating}. Recently, MOD in satellite videos has received increasing attention. Despite its significance, fast and accurate SVMOD is difficult due to at least the following challenges in realistic applications.

(1) \textbf{Real-time requirements}. As there are vast amounts of satellite videos that are temporally redundant,  of large spatial size, and used for time-critical applications, real-time SVMOD methods are important. While in satellite videos, the background is temporally redundant, the foreground is highly sparse, and dynamic information is critical,  \emph{how to make good use of these characteristics to improve the efficiency of SVMOD is essential.}

(2) \textbf{High-quality requirements}. Moving objects in satellite videos \footnote{increased temporal resolution leading to the loss of the spatial resolution to some extent compared with static satellite images} are usually tiny, dim (low local contrast between foreground and background), with little shape and texture information,  and sensitive to noise. These intrinsic characteristics increase the difficulty of learning high-quality (accurate and robust) object representations. Therefore, \emph{how to develop effective and specific frameworks tailored for SVMOD balance recall and precision is also important.}

(3) \textbf{High annotation cost}. Just due to the aforementioned characteristics of moving objects and satellite videos, manually labeling these objects requires repeated inspections to confirm moving objects, which easily leads to noisy labels and is costly. It is difficult to obtain large amounts of accurately annotated training data. Therefore, \emph{how to develop label efficient solutions for SVMOD is valuable for realistic applications.}

Existing methods can be divided into two main categories: classical model-based \cite{Zhou2013Decolor,zhang2019error,Zhang2021MovingVD,yin2021moving,ahmadi2019moving,ao2019needles,yin2021detecting, 7076585_liu, 10360211_gao, 7911235_liu}, and modern learning-based \cite{xiao2021dsfnet, xiao2023incorporating, lalonde2018clusternet, pi2022very}. In the former category, differencing methods \cite{ao2019needles, yin2021detecting} and robust PCA methods \cite{Godec, Zhou2013Decolor,zhang2019error,Zhang2021MovingVD, 7076585_liu, 10360211_gao, 7911235_liu, 10317532_Naganuma} have been widely studied over the last decade. However, methods in this category mainly rely on motion information to detect moving objects, which can be easily influenced by dynamic clutters caused by uneven platform movement and illumination changes, resulting in suboptimal performance.

\begin{figure}[t]
	\centering
	\includegraphics[width=8cm]{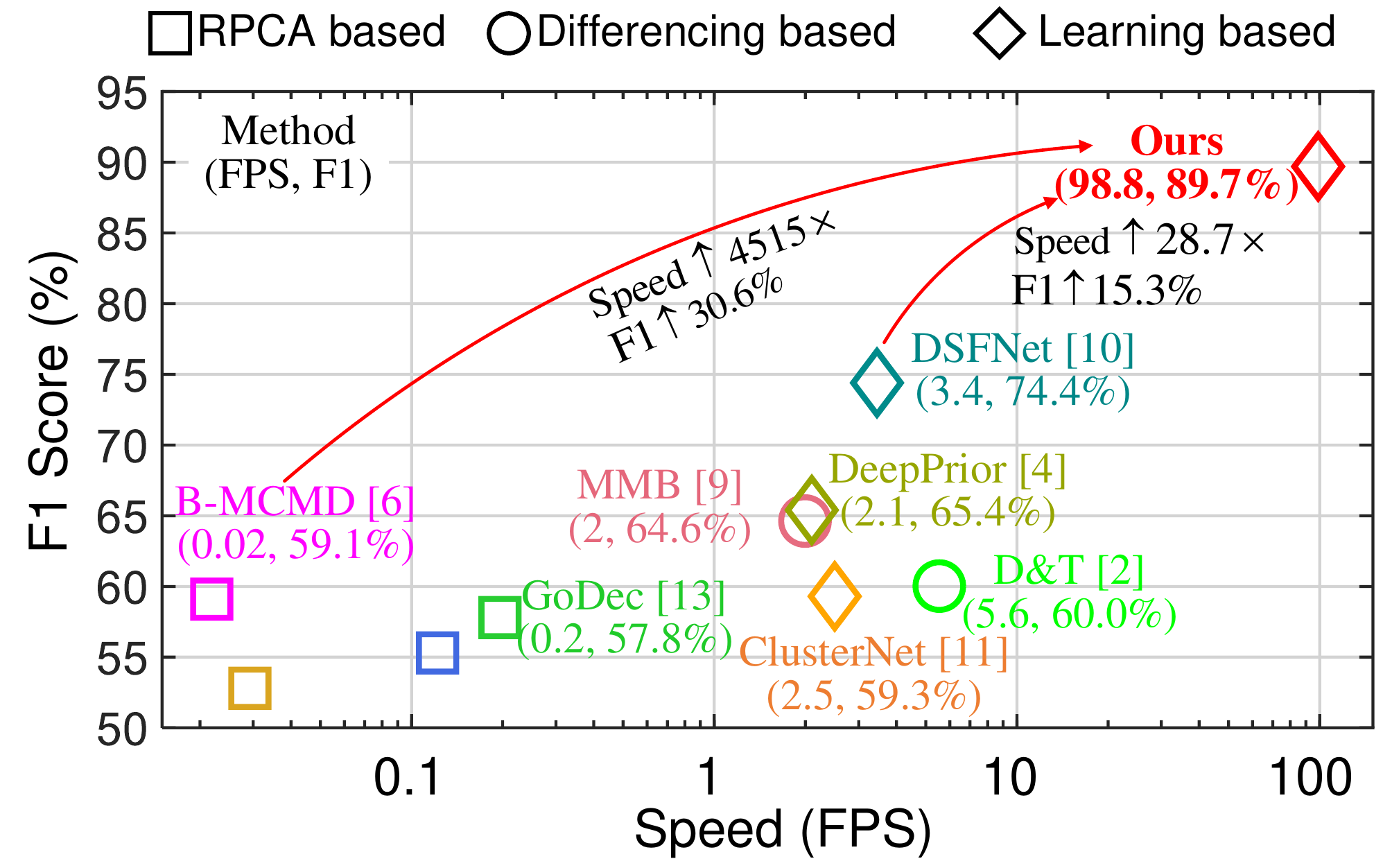}
	\caption{  Comparison of the detection performance (F1 score) and computational efficiency (Frame Per Second, FPS) of different methods on the re-labeled VISO dataset. Our proposed unsupervised method is highly efficient and effective compared with other methods, including RPCA-based (GoDec \cite{Godec}, DECOLOR \cite{Zhou2013Decolor}, E-LSD \cite{zhang2019error} and B-MCMD \cite{Zhang2021MovingVD}), differencing-based (D\&T \cite{ao2019needles} and MMB \cite{yin2021detecting}) and learning-based methods (ClusterNet \cite{lalonde2018clusternet}, DSFNet \cite{xiao2021dsfnet}, and DeepPrior \cite{xiao2023incorporating}). Specifically, compared with the traditional method B-MCMD \cite{Zhang2021MovingVD} and learning-based method DSFNet \cite{xiao2021dsfnet}, our method speeds up 4490x and 28.7x with a significant improvement on F1 score of 30.6\% and 15.3\%, respectively. All learning-based methods are conducted on a single RTX2080Ti GPU.}
	\label{convergence}
 % \vspace{-2mm}
\end{figure}

Recently, deep learning has brought promising progress for SVMOD \cite{xiao2021dsfnet, xiao2023incorporating, pi2022very}. However, unlike the extensively studied problem of generic object detection \cite{Liu2019DeepLF, Chen2020ASO} in the field of computer vision, deep learning for the problem of SVMOD remains largely underexplored, with only a few works. Representative methods focus on exploring spatio-temporal information to detect moving objects \cite{lalonde2018clusternet, xiao2021dsfnet, pi2022very}. For example, Xiao \emph{et al.} \cite{xiao2021dsfnet} explores static semantic information and dynamic motion cues to tackle SVMOD, which utilizes computation-intensive 3D convolution to extract spatio-temporal information. Pi \emph{et al.} \cite{pi2022very} utilize weighted element-wise multiplication between pair-wise differencing images to suppress dynamic clutters. Although achieving higher performance, these methods are computationally expensive as they are multi-frame-based. Moreover, deep learning \cite{xiao2023incorporating} requires vast amounts of labeled data for training that are difficult to obtain as discussed previously. Therefore, given our previous detailed analysis, the focus of this work is to develop a highly efficient (computation and label efficient) yet effective method for SVMOD. 

To this end, in this work, we deeply looked into the inherent characteristics (such as the aforementioned challenges) of the problem of SVMOD and propose a \emph{simple}, \emph{novel}, yet \emph{highly efficient} and \emph{unsupervised} framework for SVMOD, named Highly Efficient and Unsupervised Moving object (vehicle) detection (HiEUM). Our proposed HiEUM framework is based
on two core designs. First, we develop a label self-evolution unsupervised training framework to reduce the cost of annotation and study the possibility of unsupervised SVMOD. Specifically, we utilize a traditional method to generate initial pseudo labels and update these coarse labels by inferencing the proposed model during training. Second, we design a highly efficient sparse convolutional anchor-free detection network by formulating sparse sampling coarse target regions as spatio-temporal point clouds, which is motivated by the extremely sparse nature of moving objects, temporal dynamic information of moving objects, and high redundancy of background in satellite videos. Coupling these two novel designs enables us to establish a highly efficient and effective framework for SVMOD. 
 
Our experiments (as shown in Figure~\ref{convergence}) show that we accelerate the inference significantly (nearly $28.7\times$ compared with SOTA DSFNet \cite{xiao2021dsfnet}, $4490\times$ compared with traditional B-MCMD published in PAMI \cite{Zhang2021MovingVD}) and improve accuracy also significantly. We believe our proposed HiEUM sets the new SOTA and opens a promising direction for SVMOD.

Our contributions to this work are summarized as follows.
\begin{itemize}

	\item We introduce a label self-evolution unsupervised framework in which pseudo-labels can evolve with the training process that is well compatible with existing learning-based detection methods for SVMOD.
 
 	\item We propose a sparse convolutional anchor-free detection network for SVMOD, which exploits the sparsity of moving targets to skip redundant computation on background regions. Unlike existing methods, our method is the first attempt to handle SVMOD via a sparse spatio-temporal point cloud representation.
 
	\item To verify the detection performance of small and dim moving targets in satellite videos, we propose a new SVMOD dataset with re-labeled dim and small moving vehicles from the VISO dataset. We also provide a new benchmark with various methods under the newly labeled dataset.
\end{itemize}

The rest of this paper is organized as follows. Some related works are reviewed in Section \ref{relatedwork}. The proposed framework is described in Section \ref{proposedmethod}. Section \ref{experiments} presents the experimental setup and results in detail. Section \ref{conclusion} concludes this paper.

% \vspace{-5mm}
\section{Related Work}
\label{relatedwork}

\subsection{Methods for SVMOD}
% \textbf{(1) Methods for SVMOD} 

Currently, methods for SVMOD can be divided into model-based methods and learning-based methods.

Model-based methods mainly exploit the prior knowledge of the targets to obtain SVMOD results, including frame differencing-based methods \cite{Saleemi2013MultiframeMP, Keck2013RealtimeTO} and background subtraction-based methods \cite{Liu2019DeepLF, Jiao2021NewGD, ren2016faster, zhou2019objects}. The former exploits two- or three-frame differencing methods to extract moving vehicles, and the latter reconstructs the background to obtain moving foreground through background subtraction. However, the model-based methods rely on hand-crafted features with complex parameter tuning processes and can not handle severe scenario changes with fixed hyper-parameters.

Recently, with the powerful modeling ability, learning-based methods have improved the detection performance of both general object detection \cite{cheng2023towards,  Liu2019DeepLF} and moving vehicles in satellite videos \cite{lalonde2018clusternet, xiao2021dsfnet, pi2022very, feng2023SDANet}. Lalonde et al. \cite{lalonde2018clusternet} proposed a two-stage network named ClusterNet to detect small objects in airborne images. To handle images with large sizes, they first extracted regions of interest in the first stage and then got the fine-grained detection results in the second stage. Xiao et al. \cite{xiao2021dsfnet} proposed a two-stream network to combine static context information from a single image and dynamic motion cues from multiple images, which significantly promoted the detection performance but with limited efficiency due to the intensive computation of multiple frames. Pi et al. \cite{pi2022very} proposed to integrate motion information from adjacent frames through frame differencing and exploit transformer to refine the extracted features for better discrimination ability of actual moving targets. Feng et al. \cite{feng2023SDANet} proposed a semantic-embedded density adaptive network named SDANet to incorporate the semantic features of the road to emphasize the regions of interest. In conclusion, learning-based methods have been demonstrated to be effective for SVMOD. However, the existing learning-based methods process images in dense representation and multi-frame form, hindering them from balancing efficiency and effectiveness and extracting long-term spatio-temporal information.

In this paper, we explore the sparsity of moving vehicles in satellite videos and process dozens of images in sparse point cloud representation to extract long-term information for better detection performance and skip uninformative background regions for higher inference speed.

\subsection{Unsupervised Learning for SVMOD}
% \textbf{(2) Unsupervised Learning for SVMOD}

Supervised learning-based methods are restricted to a large amount of annotated data. Semi-supervised learning-based methods can help to alleviate the burden of annotation costs \cite{9288631_Giraldo}. For example, Giraldo et al. \cite{9288631_Giraldo} propose a graph learning-based method under the semi-supervised framework to tackle moving object segmentation, in which the concepts of graph signal processing \cite{10146241_Leus} are introduced.

Unsupervised learning is widely used in many fields \cite{bao2022discovering,Gao_Large} due to the independence of manual annotations. Currently, many unsupervised methods have been proposed for moving object detection in natural images \cite{zhuo2019unsupervised,bao2022discovering}. However, these methods can not be directly adopted for SVMOD due to the extremely dim and small target characteristics.

To ease the time-consuming annotation process of moving objects in satellite videos, Zhang et al. \cite{zhang2021learning} proposed a weakly supervised method where the supervision signal is produced by a traditional method E-LSD \cite{zhang2019error}. Due to the inaccuracy of the generated labels, the performance of their method is inferior to the original label generation method \cite{zhang2021learning}. Xiao et al. \cite{xiao2023incorporating} combined an unsupervised background reconstruction network with traditional iterative optimization to detect moving vehicles in satellite videos. However, the detection performance and efficiency need to be further improved.

In this paper, we propose an unsupervised framework to improve the accuracy of pseudo-labels produced by traditional methods. We propose to promote the quality of pseudo labels by utilizing the consistency of object trajectories and iteratively updating the generated labels during training.

\subsection{The Priors of SVMOD}
% \textbf{(3) The Priors of SVMOD}

Different from general object detection, moving vehicle detection has its characteristics, such as the small sizes and the consistency of movement. Traditional model-based methods exploit these priors to distinguish small moving targets from satellite videos. The most usually used priors are summarized as follows:

\textbf{The continuity of movement}. Due to the consistent movement of the vehicles in satellite videos, the trajectory of the moving target is a prominent characteristic of the targets, which can be used to filter out false alarms \cite{ao2019needles, ahmadi2019moving, yin2021detecting}.

\textbf{The sparsity of the targets}. Due to the tiny sizes of the vehicles, the moving targets occupy only a small ratio of the whole image, which can be considered as sparse compared to the large image sizes \cite{zhang2019error, Zhang2021MovingVD, Zhou2013Decolor, Godec}. 

\textbf{The low-rank property of the background}. Although satellite platform motion exists between adjacent frames, the background in consecutive frames overlaps to a large extent. Therefore, The matrix or tensor of background from multiple frames can be considered as low-rank \cite{zhang2019error, Zhang2021MovingVD, Zhou2013Decolor, Godec, xiao2023incorporating}.

In this paper, we extensively explore combining these priors into the design of learning-based methods. We get the spatio-temporal point cloud by utilizing the low-rank prior of the background and the sparsity of targets to design our sparse detection network, which can skip redundant computation of background and significantly improve detection efficiency. 

\begin{figure*}[t]
	\centering
	\includegraphics[width=\linewidth]{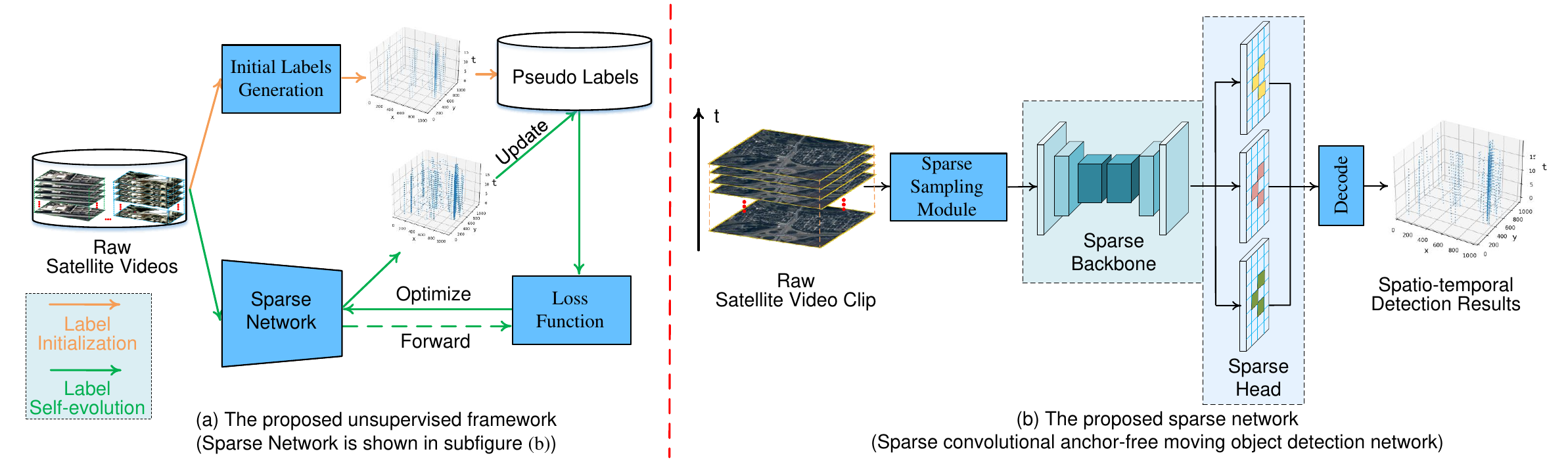}
	\caption{An illustration of our proposed unsupervised framework. (a) The overall architecture of the proposed iterative updating unsupervised framework. (b) The proposed sparse convolutional anchor-free moving object detection network. Firstly, we utilize the traditional method modified from \cite{xiao2023incorporating} and SORT \cite{bewley2016simple} to obtain initial pseudo labels. Then, the initial labels and videos are exploited to train the sparse detection network. The input video clip is processed by a sparse sampling module to extract sparse 3D spatial-temporal point cloud data. Finally, the trained sparse network is used to update the pseudo labels and retrained to promote the detection performance. Note that the method generating initial labels and the detection network can be replaced by an arbitrary method, demonstrating the flexibility of our proposed framework.}
	\label{fig:unsupframework}
 % \vspace{-5mm}
\end{figure*}

% \vspace{-3mm}
\section{Proposed Methodology}
\label{proposedmethod}

In this section, we first illustrate our label self-evolution unsupervised framework and then present the sparse convolutional anchor-free moving object detection network in detail.

% \vspace{-3mm}
\subsection{Label Self-evolution Unsupervised Framework} \label{unsupframework}

Due to the small size and low contrast to the background, moving vehicles in satellite videos are difficult to distinguish from the heavy clutters, thus improving the annotation complexity. To relieve the burden of annotating large-scale datasets for SVMOD, we proposed a label-evolution unsupervised framework that utilizes traditional methods for the initialization of labels and updates the labels during training to promote detection performance. The proposed unsupervised framework is shown in Fig. \ref{fig:unsupframework} (a).  

The proposed framework first generates pseudo labels by using a traditional method. Here, we utilize the method in \cite{xiao2023incorporating} and replace the background reconstruction network with a simple temporal median filter to obtain the background in a fast way. It is worth noting that the proposed framework is flexible and can be applied by using any traditional detection methods. However, due to the limited detection performance of traditional methods, the initial pseudo labels contain some false alarms and miss some dim targets, thus limiting the detection performance of learning-based methods.

Therefore, we utilize two strategies to improve the quality of the pseudo labels. On the one hand, to alleviate the influence of false alarms, we exploit SORT \cite{bewley2016simple} to get the trajectory of moving objects and utilize trajectory length and velocity constraints to filter out false alarms. On the other hand, to recover the missed dim targets, we iteratively update the initial labels during the training process. Specifically, when the network is trained for every 10 epochs, we use the trained network to infer on the train set to generate new pseudo-labels. To reduce false alarms in these new pseudo-labels, we also apply SORT \cite{bewley2016simple} to filter out false alarms. To prevent the network from over-fitting on the self-generated labels, we retain the initial pseudo-labels and add new labels generated by the trained network as the new train set. Through continuous iteration, the label quality can be improved, thereby improving the detection performance of the learning-based methods.

In conclusion, the proposed framework is generic and flexible, in which the traditional and learning-based method can be replaced by arbitrary properly designed methods. Moreover, this paper proves that the spatio-temporal consistency prior of moving objects (i.e., the consistency of the trajectory of targets) can help learning-based methods achieve good performance without any manual annotation.

% \vspace{-3mm}
\subsection{Sparse Convolutional Anchor-free Moving Object Detection Network} \label{sparse_network}

Due to the extreme imbalance between foreground and background regions, most computational resources of learning-based methods are assigned to uninformative background areas, which incurs tremendous redundant computational burdens and hinders the extraction of long-term spatio-temporal information.  

Motivated by the sparsity of moving targets in satellite videos, we propose a sparse convolutional anchor-free moving object detection network to exploit the sparsity prior for both effective and efficient SVMOD. As shown in Fig. \ref{fig:unsupframework} (b), Our proposed network consists of three parts, i.e., the sparse sampling module, the sparse backbone, and the sparse detection head. 

\begin{figure}[t]
	\centering
	\includegraphics[width=7cm]{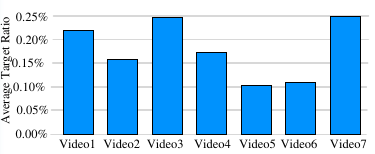}
	\caption{The average target ratio of the test set. The average ratio of the target in each video is less than 0.3\%, which demonstrates the sparsity of the moving vehicle in satellite videos.}
	\label{distribution}
\end{figure}

\textbf{(1) Sparse Sampling Module}: Due to the extremely small sizes, the moving vehicles in satellite videos occupy only a small portion of the whole image, which can be considered intrinsically sparse \cite{zhang2019error, zhang2020online, Zhang2021MovingVD, yin2021moving}. Figure. \ref{distribution} illustrates the average target ratio of the test set, which demonstrates that the target is sparse and the background area dominates the whole image.

\begin{figure}[t]
	\centering
	\includegraphics[width=\linewidth]{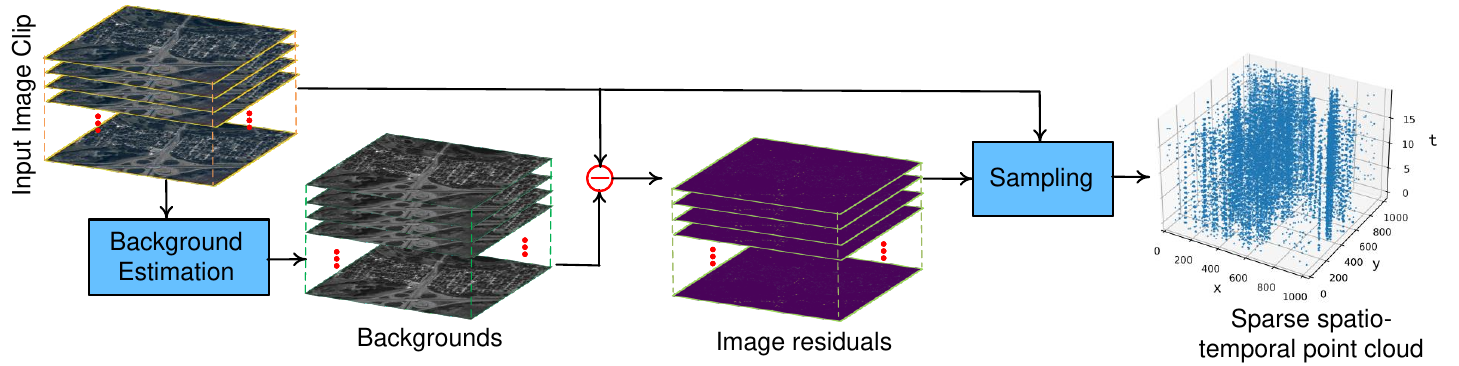}
	\caption{The illustration of the proposed sparse sampling module.}
	\label{sparse_sampling_module}
 % \vspace{-5mm}
\end{figure}

To reduce the redundant background regions in satellite videos, we design a sparse sampling module that exploits frame-differencing to subtract background regions. The procedure is shown in Fig. \ref{sparse_sampling_module}. We first estimate the background using a temporal median filter. Then, the residual images are generated by subtracting the estimated background from the original image sequence. In this way, the background can be significantly reduced. However, many small residuals still exist in the background regions. To further alleviate the influence of background regions, we leverage an adaptive threshold to segment the candidate target regions.

We utilize an adaptive threshold with the parameter $k$ to produce a proper threshold, which can be obtained by
\begin{equation}
	th = \mu  + k \times \sigma,
\end{equation}
where $\mu$ and $\sigma$ denote the mean and standard deviation of the residual image, respectively. $k$ is a predefined value that controls the maximum threshold. 

Note that the threshold has a great impact on the detection performance. When the threshold is too small, more points would be assigned to the candidate target region, which would increase the computational cost. In contrast, when the threshold is too large, not all the target points can be correctly picked out, which would damage the detection performance. Therefore, the threshold should be carefully selected to balance accuracy and efficiency. 

After the segmentation of the adaptive threshold, we can extract coarse foreground regions from multiple images. Although most of the background is removed, regular learning-based methods still process all locations equally, which incurs a lot of redundant computation. To reduce redundant computations, we extract the valid foreground pixels and reshape them as a sparse spatio-temporal 3D point cloud.

\begin{figure}[t]
	\centering
        \setlength{\belowcaptionskip}{-5mm}
	\includegraphics[width=8cm]{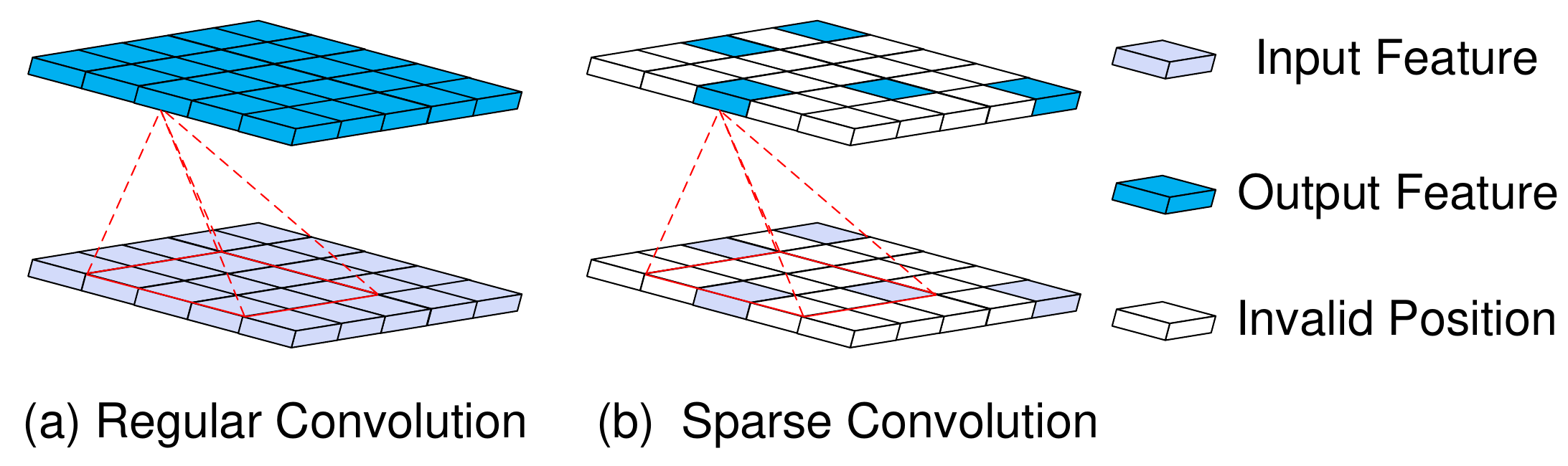}
	\caption{An illustration of sparse convolution. Compared with regular convolution, sparse convolution only calculates on valid positions, \emph{i.e.} the blue square.}
	\label{sparse_conv}
 % \vspace{-3mm}
\end{figure}

\textbf{(2) Backbone}: To cope with the sparse data structure and skip background regions, we build our detection network based on sparse convolution (as shown in Fig. \ref{sparse_conv}), which is widely used in point cloud processing. Due to the extremely small sizes of moving vehicles in satellite videos, down-sampling operations could lead to information loss of small moving targets. To tackle this issue and preserve the feature of small moving targets, a U-net structure network \cite{shi2020points} composed of sparse convolution is employed as the backbone to extract features from the sparse spatio-temporal point cloud. Note that the backbone can be substituted by any other network that can process point clouds, which indicates the flexibility of our proposed detection network. Besides, compared with the original dense image representation and conventional 3D convolution-based methods \cite{xiao2021dsfnet}, our proposed method can significantly reduce computational requirements, which holds immense potential for handling massive data from large-scale remote sensing images. 

\textbf{(3) Sparse Head}: Different from anchor-based methods which would generate tremendous anchor boxes in our case, we design a sparse anchor-free head to predict the locations of the moving target. Inspired by CenterNet \cite{zhou2019objects}, our designed head consists of three parallel sparse convolutional branches to predict the object centers, sizes and offsets, respectively. To conveniently compute the loss and decode the detection results, we transform the results from the sparse form into dense representations. Note that, different from previous multi-to-one detection methods (i.e., DSFNet \cite{xiao2021dsfnet}, ClusterNet \cite{lalonde2018clusternet} and VLR-MVD \cite{pi2022very}), our proposed method is designed on a multi-to-multi paradigm, which can further enhance the efficiency.

In the training phase, we use center loss, size prediction loss, and offset loss as used in \cite{zhou2019objects} to guide the network to learn to detect moving targets. The overall loss is formulated as follows:
$L = L_{ctr}+\lambda _1 L_{size} + \lambda _2 L_{off}$ 
where $L_{ctr}$, $L_{size}$, and $L_{off}$ are the center loss, size prediction loss, and offset loss, respectively. $\lambda _1$ and $\lambda _2$ denote the penalty factor. In the testing phase, we parallelly decode the results from the head to obtain moving objects of multiple frames simultaneously, which could further improve the efficiency of the whole detection process.

\textbf{(4) Discussion}: Previous methods \cite{xiao2021dsfnet, pi2022very} detect moving targets directly from the original dense representation, which contains much redundant computation, thus limiting the real-time employment and the long-term information modeling on the time domain. Due to the low-rank characteristic of the background region and the intrinsic sparsity of the moving targets in satellite videos, the redundant regions can be roughly removed through background subtraction and the residuals can be filtered to construct a sparse 3D point cloud. Since the 3D point cloud has much fewer pixels than the original multi-frame images, more frames can be processed simultaneously and the long-term spatio-temporal modeling can be achieved at a low cost on the hardware and computational resources. Therefore, our method has two remarkable advantages.

\textbf{Highly efficient on SVMOD}. Most existing methods generate detection results from the raw images while we process the images from a new perspective (i.e., the spatio-temporal 3D point cloud), which can reduce tremendous redundant computation and thus can improve efficiency a lot.

\textbf{Highly effective on SVMOD}. High efficiency always comes at the cost of detection performance degradation. However, our method can achieve both efficiency and effectiveness, which is benefited from the long-term spatio-temporal modeling as well as the removal of clutter interferences.

However, everything comes at a cost. Although the redundant background region can be reduced, the target regions can also be wrongly removed due to the imperfect background modeling, especially for large moving targets and extremely dim targets, which can not be recovered in the subsequent steps. One can further improve the detection performance with fine-grained background modeling. But note that complex background modeling could incur additional computational burdens. Therefore, we should consider the balance between efficiency and effectiveness when designing background modeling methods.

% \vspace{1cm}
\begin{table*}[t]
	\centering
	\renewcommand\arraystretch{1.1}
	\footnotesize
	\caption{Detection performance achieved by different methods (i.e., GoDec \cite{Godec}, DECOLOR \cite{Zhou2013Decolor}, E-LSD \cite{zhang2019error}, B-MCMD \cite{Zhang2021MovingVD}, D\&T \cite{ao2019needles}, MMB \cite{yin2021detecting}, ClusterNet \cite{lalonde2018clusternet}, DSFNet \cite{xiao2021dsfnet}, DeepPrior \cite{xiao2023incorporating} and the proposed HiEUM). The best results are shown in \textbf{boldface} and the second best results are shown in \underline{underline}.}\label{results_table1}
	\setlength{\tabcolsep}{1mm}{
		\resizebox{\linewidth}{!}{
			\begin{tabular}{c|ccc|ccc|ccc|ccc|ccc|ccc|ccc|ccc}
				\midrule[0.75pt]
				\multirow{2}*{Method} & \multicolumn {3}{c}{Video1}  &  \multicolumn {3}{c}{Video2}  & \multicolumn {3}{c}{Video3}  &  \multicolumn {3}{c}{Video4} &\multicolumn {3}{c}{Video5} &\multicolumn {3}{c}{Video6} 
				&\multicolumn {3}{c}{Video7}
				&\multicolumn {3}{c}{AVERAGE} \\
				\cmidrule(r){2-4} \cmidrule(r){5-7} \cmidrule(r){8-10} \cmidrule(r){11-13}	\cmidrule(r){14-16} \cmidrule(r){17-19} \cmidrule(r){20-22} \cmidrule(r){23-25}
				&Re&Pr&F1  &Re&Pr&F1  &Re&Pr&F1  &  Re&Pr&F1   &Re&Pr&F1 &Re&Pr&F1 	&Re&Pr&F1  &Re&Pr&F1\\\midrule[0.75pt]
				GoDec \cite{Godec}
				&70.5	&79.5	&74.8
                &50.5	&86.1	&63.7
                &37.7	&87.7	&52.7
                &51.9	&81.3	&63.4
                &45.9	&76.5	&57.4
                &45.4	&74.8	&56.5
                &24.4	&68.2	&35.9
                &46.6	&79.2	&57.8
				% &5.10
				\\
				DECOLOR \cite{Zhou2013Decolor}
                &29.4	&\textbf{99.7}	&45.5
                &52.8	&90.4	&66.6
                &45.9	&86.9	&60.1
                &34.2	&\textbf{98.9}	&50.9
                &51.5	&88.8	&65.2
                &\underline{55.5}	&77.5	&64.7
                &21.1	&89.1	&34.1
                &41.5	&90.2	&55.3
				% &8.20
				\\
				E-LSD \cite{zhang2019error}
                &58.8	&77.4	&66.8
                &40.5	&41.4	&40.9
                &36.2	&92.4	&52.0
                &46.5	&90.2	&61.3
                &35.6	&87.3	&50.6
                &34.0	&81.8	&48.1
                &37.8	&74.8	&50.2
                &41.3	&77.9	&52.8
				% &34.20
				\\
				B-MCMD \cite{Zhang2021MovingVD}
                &62.8	&94.4	&75.5
                &46.2	&88.0	&60.6
                &40.2	&83.0	&54.2
                &56.4	&72.2	&63.3
                &31.4	&81.6	&45.4
                &42.7	&74.1	&54.2
                &58.9	&62.2	&60.5
                &48.4	&79.4	&59.1
				% &45.70
				\\
				\hline
				D\&T \cite{ao2019needles}
                &59.3	&93.1	&72.4
                &43.0	&89.8	&58.2
                &40.1	&87.3	&54.9
                &59.4	&83.2	&69.3
                &36.7	&85.8	&51.4
                &41.1	&81.2	&54.6
                &\underline{60.9}	&56.9	&58.8
                &48.6	&82.5	&60.0
				% &\underline{0.18}
				\\
				MMB \cite{yin2021detecting}
                &66.5	&95.9	&78.6
                &45.3	&94.6	&61.2
                &40.9	&94.7	&57.2
                &63.0	&86.8	&73.0
                &40.1	&90.3	&55.5
                &42.3	&91.4	&57.9
                &56.5	&87.5	&68.6
                &50.7	&91.6	&64.6
				% &0.50
				\\
				\hline
				ClusterNet \cite{lalonde2018clusternet}
                &61.9	&65.1	&63.4
                &41.7	&86.8	&56.3
                &44.9	&78.8	&57.2
                &41.1	&74.2	&52.9
                &43.7	&85.7	&57.9
                &51.4	&80.8	&62.8
                &53.4	&82.2	&64.7
                &48.3	&79.1	&59.3
				% &0.40
				\\
				DSFNet\cite{xiao2021dsfnet}
                &\underline{76.9}	&94.9	&\underline{85.0}
                &\underline{58.5}	&\underline{96.3}	&\underline{72.8}
                &\underline{50.5}	&94.6	&\underline{65.9}
                &\underline{70.3}	&\underline{98.2}	&\underline{81.9}
                &\underline{73.9}	&\underline{95.4}	&\underline{83.3}
                &53.9	&\textbf{96.2}	&\underline{69.1}
                &46.3	&\textbf{98.0}	&62.9
                &\underline{61.5}	&\textbf{96.2}	&74.4
				% &0.29
				\\
                \hline
                DeepPrior\cite{xiao2023incorporating}
                &73.2	&84.8	&78.6
                &48.3	&\underline{96.3}	&64.3
                &43.6	&\underline{96.3}	&60.0
                &55.0	&92.6	&69.0
                &40.4	&94.3	&56.6
                &46.0	&94.6	&61.9
                &53.6	&90.1	&\underline{67.2}
                &51.4	&92.7	&65.4
                % &0.40
                \\
                HiEUM(Ours)
                &\textbf{85.5}	&\underline{97.3}	&\textbf{91.1}
                &\textbf{82.2}	&\textbf{96.8}	&\textbf{88.9}
                &\textbf{78.1}	&\textbf{96.6}	&\textbf{86.3}
                &\textbf{92.1}	&95.7	&93.9
                &\textbf{82.2}	&\textbf{97.7}	&\textbf{89.3}
                &\textbf{84.5}	&\underline{95.8}	&\textbf{89.8}
                &\textbf{84.7}	&\underline{92.8}	&\textbf{88.5}
                &\textbf{84.2}	&\underline{96.1}	&\textbf{89.7}
                \\
				\midrule[0.75pt]
	\end{tabular}}}
  % \vspace{-3mm}
\end{table*}

\begin{figure}[t]
	\centering
        \setlength{\belowcaptionskip}{-5mm}
	\includegraphics[width=8cm]{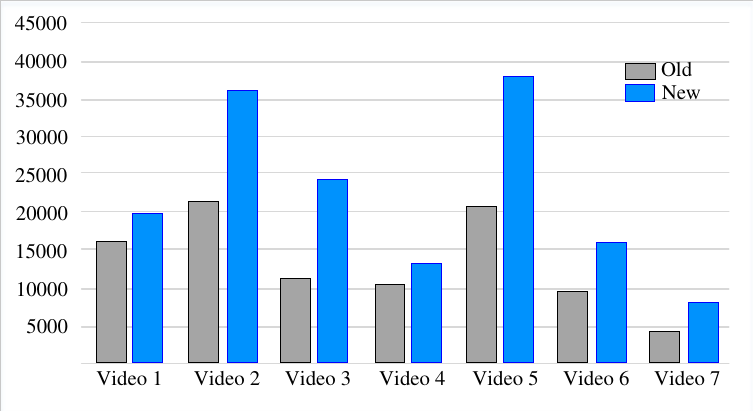}
	\caption{The comparison of old and new labeled target numbers. After re-labeling, the number of annotations increases sharply, demonstrating that each video contains many dim and small moving targets that are ignored in the previous dataset. }
	\label{data_num}
\end{figure}

% \vspace{-6mm}
\section{Experimental Results and Discussions}
\label{experiments}  

\subsection{Dataset Description}
The detection performance of the proposed method is evaluated on satellite videos from Jilin-1 satellite. The ground sampling distance (GSD) of the dataset is 0.92 meters, and the frame rate is 10 frames per second. The moving vehicles in the dataset are labeled by bounding boxes as the ground truth.  

\begin{figure}[t]
	\centering
	\includegraphics[width=1 \linewidth]{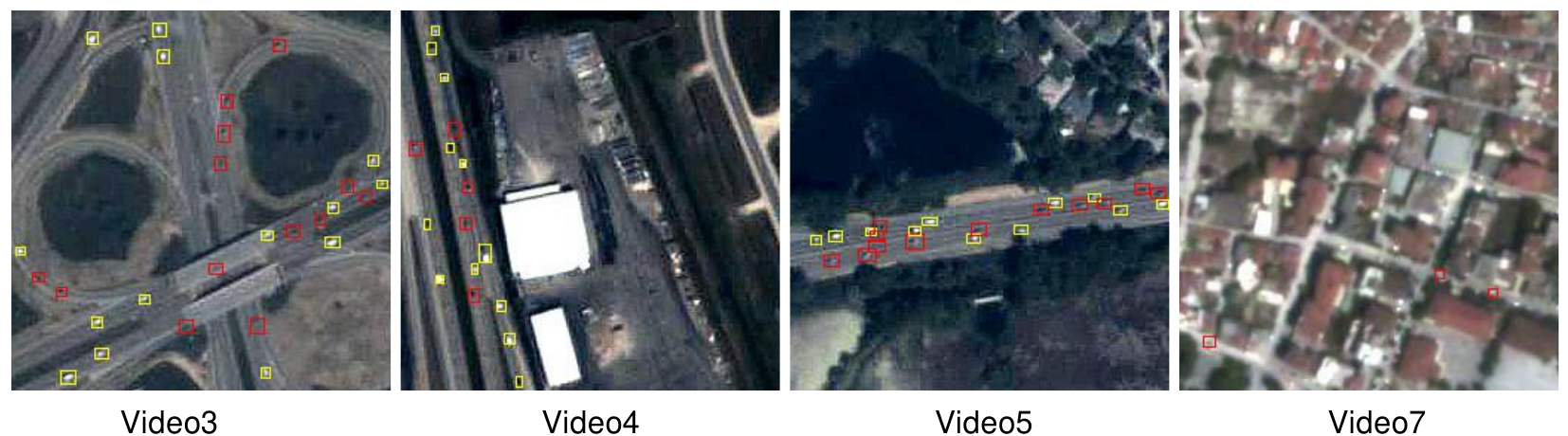}
	\caption{Qualitative results of new and old labels. Rectangles in yellow and red represent the original and newly labeled targets.}
	\label{relabels}
 % \vspace{-2mm}
\end{figure}

In the previously released dataset, most dim targets are ignored for annotation, which would incur an unfair comparison of different methods. To address this issue, we relabel the test set to include dim and small targets for a fair comparison. Through relabeling, we find that the former dataset missed a lot of dim targets. The comparison of target numbers of the former and relabeled datasets are shown in Fig. \ref{data_num}. It can be observed that the target number of all the videos in the test set has significantly increased after relabeling. The relabeled test set contains 155987 instances, while the former test set contains only 93491 instances. Moreover, we have rechecked the old labels and adjusted some labels that are of poor quality. This new relabeled dataset can serve as a new benchmark for dim and moving vehicle detection. Figure. \ref{relabels} shows the annotations of several typical scenarios before and after relabeling. It can be observed that many dim targets (in red rectangles) are relabeled to evaluate the dim moving target detection performance of different methods.

\subsection{Implementation Details and Evaluation Criteria}

We used 20 consecutive frames as input. The batch size was set to 6 with a random crop image patch size of $256 \times 256$. We trained our network using the Adam optimizer \cite{Kingma2015AdamAM} for 55 epochs with an initial learning rate of $1.25 \time 10^{-4}$. The learning rate was decreased by a factor of 10 after 30 and 45 epochs. To avoid over-fitting on the train set, we sampled 1/5 of the train set for training and the whole test set for evaluation. To remove false alarms in the initial pseudo labels, we filtered out the trajectory length of less than 30 points and the mean velocity of targets of less than 0.55 pixels per frame. We iteratively updated the pseudo labels every 10 epochs. All the models were implemented with Pytorch on two Nvidia RTX 2080Ti GPUs.

In this paper, we follow \cite{lalonde2018clusternet, xiao2023incorporating, zhang2019error, Zhang2021MovingVD} to use precision (Pr), recall (Re), and F1 score (F1) as the evaluation metrics. The average precision (avg Pr), average recall (avg Re), and average F1 score (avg F1) on all the test videos are also used for evaluation. We follow \cite{lalonde2018clusternet, xiao2023incorporating} to employ distance metric to determine whether a detected result is a positive sample. The distance threshold is set to 5 pixels for performance evaluation. 

\subsection{Comparison to the State-of-the-arts}
In this subsection, we present the detection results and analyses of different SVMOD methods. We compare the proposed method with several state-of-the-art methods, including 2 frame-differecing based methods (i.e., D\&T \cite{ao2019needles} and MMB \cite{yin2021detecting}), 4 RPCA-based methods (i.e., GoDec \cite{Godec}, DECOLOR \cite{Zhou2013Decolor}, E-LSD \cite{zhang2019error} and B-MCMD \cite{Zhang2021MovingVD}), 2 supervised learning-based methods (i.e., ClusterNet \cite{lalonde2018clusternet} and DSFNet \cite{xiao2021dsfnet}), and 1 unsupervised learning-based method (DeepPrior \cite{xiao2023incorporating}). 

\textbf{(1) Quantitative Results}. The quantitative results are shown in Table \ref{results_table1}. It can be observed that our proposed HiEUM can achieve the best detection performance with the highest average F1 score of 89.7\%, outperforming the second-best method DSFNet with 15.3\%. We attribute this to two reasons. On the one hand, we have iteratively updated the training labels, which can evolve for better quality labels and include unlabeled dim targets. On the other hand, due to the low memory occupancy and computational burdens, our HiEUM can process more frames for long-term spatio-temporal information to promote detection performance. For example, our HiEUM can process 20 frames to predict all the results of the input frames at a time, while DSFNet \cite{xiao2021dsfnet}) can only process 5 frames for the results of one frame. 

\begin{figure*}[t]
	\centering
	\includegraphics[width=\linewidth]{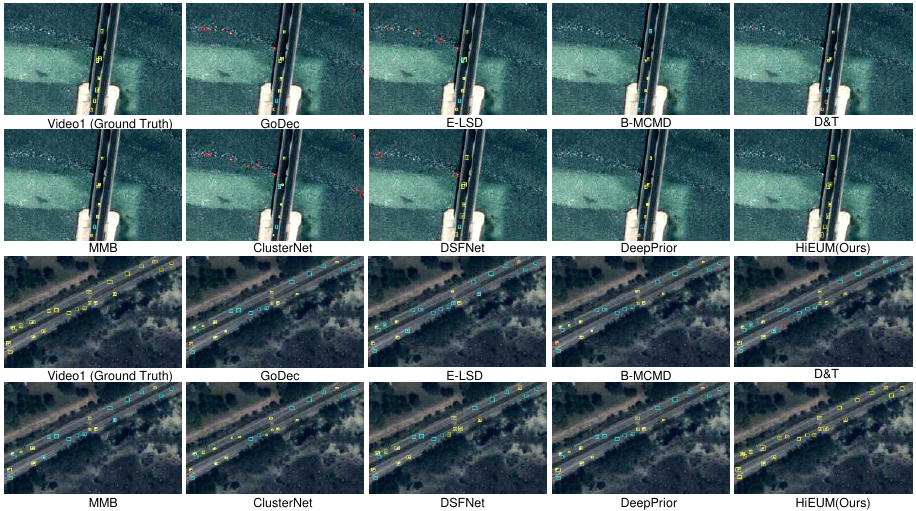}
	\caption{Qualitative results of different methods (i.e., GoDec \cite{Godec}, E-LSD \cite{zhang2019error}, B-MCMD \cite{Zhang2021MovingVD}, D\&T \cite{ao2019needles}, MMB \cite{yin2021detecting}, ClusterNet \cite{lalonde2018clusternet}, DSFNet \cite{xiao2021dsfnet}, DeepPrior \cite{xiao2023incorporating} and the proposed HiEUM). For each scene, the red, green, and yellow boxes represent the false, miss, and correct detections, respectively.}
	\label{quali_results}
  % \vspace{-3mm}
\end{figure*}

\begin{table}
	% \vspace{-.05in}
	\centering
	\renewcommand\arraystretch{1.1}
	\caption{The performance of different detection methods. The best results are shown in \textbf{boldface}.}\label{supervised_mode}
	{
		\begin{tabular}{c|c|c|c}
			\hline
			methods &Avg Re(\%) &Avg Pr(\%) &Avg F1(\%)\\[0.05cm]
			\hline
   			DSFNet    
                &61.5	&96.2	&74.4
                \\
			HiEUM-sup    
                &71.0	&\textbf{98.2}	&82.1
                \\
			HiEUM-unsup %w/ SORT    
                &60.3	&97.5	&74.1
                \\
                % unsup w/o SORT    
                % &0.4	&21.6	&0.8
                % \\
			HiEUM  
                &\textbf{84.2}	&96.1	&\textbf{89.7}
                \\
			\hline
	\end{tabular}}
  % \vspace{-3mm}
\end{table}

\textbf{(2) Qualitative Results}. The qualitative results of different methods are shown in Fig. \ref{quali_results}. It can be observed that our proposed method has detected all the moving targets in both scenes while the comparative methods have more or less missed targets and false detections. Note that the improvement of our method is more significant when handling dim targets (such as video 6 with many dim targets). We attribute this to long-term spatio-temporal information modeling, which can benefit the detection of dim and small moving targets in satellite videos. Moreover, our method is more robust to the dynamic clutters in the scene (such as video 1 with dynamic illumination changes), which is also benefited from the long-run spatio-temporal information. 

\textbf{(3) Time Efficiency Analysis}. As shown in Fig. \ref{convergence}, our HiEUM can process images with the size of $1024 \times 1024$ at the speed of 98.8 frames per second, while the fastest comparative method can only run at 5.6 frames per second. That is because our method only processes the valid candidate regions and skips the redundant computation of the background regions. 
 % \vspace{-3mm}
\subsection{Ablation Study}

In this subsection, we conduct different ablation studies to investigate the design of HiEUM.

\textbf{(1) Effectiveness of the Unsupervised Framework.} 
 We use DSFNet \cite{xiao2021dsfnet} as the baseline and train our sparse network under manual annotations, labels produced by traditional methods, and our proposed unsupervised framework, respectively. The quantitative results are shown in Table. \ref{supervised_mode}. It can be observed that, under the same supervision of manual labels, our HiEUM-sup outperforms DSFNet \cite{xiao2021dsfnet} by 7.7\% in terms of average F1 score due to the long-term spatio-temporal modeling. Moreover, under the supervision of labels produced by the traditional method, our HiEUM-unsup suffers a performance degradation of 8.0\% compared with HiEUM-sup in terms of average F1 score. This is because the labels produced by the traditional method can not cover all the objects in the scene, thus leading to insufficient learning of moving objects. It is worth noting that HiEUM-unsup achieves comparable results to DSFNet \cite{xiao2021dsfnet}, which demonstrates the effectiveness of our sparse network. By iteratively updating the labels, our HiEUM can greatly improve the detection performance due to the evolution of the labels.

\textbf{(2) Effectiveness of Frame Differencing.} Since the results of frame differencing can be directly used to obtain detection results via an adaptive threshold, we compare our sparse network with the method with only an adaptive threshold. As shown in Table \ref{minus}, our HiEUM outperforms the adaptive threshold by a large margin (F1 score 89.7\% versus 66.1\%), which demonstrates the effectiveness of our proposed HiEUM. Furthermore, since some of the targets are too dim to be submerged in the background, we obtain the absolute values of the residuals generated by frame differencing before the adaptive threshold to get the dim targets. The effectiveness of absolute value operation is also shown in Table \ref{minus}. It can be observed that, after taking absolute value operation, the detection performance of our HiEUM has been greatly improved due to the inclusion of the dim targets.

\begin{table}
	% \vspace{-.05in}
	\centering
	\renewcommand\arraystretch{1.1}
	\caption{The performance of different processing methods on the residuals. The best results are shown in \textbf{boldface}.}\label{minus}
	{
		\begin{tabular}{c|c|c|c}
			\hline
			methods &Avg Re(\%) &Avg Pr(\%) &Avg F1(\%)\\[0.05cm]
			\hline
                adaptive threshold
                &53.7	&89.2	&66.1
                \\
			HiEUM w/o abs   
                &62.0	&\textbf{96.5}	&75.0
                \\        
                HiEUM  
                &\textbf{84.2}	&96.1	&\textbf{89.7}
                \\
			\hline
	\end{tabular}}
  % \vspace{-3mm}
\end{table}

\textbf{(3) The Impact of Thresholds.} In our HiEUM, the adaptive threshold is a key component that controls the sampling ratio of the input frames. Specifically, the value of $k$ controls the sampling ratio of the input frames. Therefore, we investigate the influence of value $k$ on the detection performance. The quantitative results are shown in Table. \ref{thresholds}. It can be observed that, with the value of $k$ increases, the proportion of sampling points decreases, resulting in the increase of FPS.  When $k$ increases from 1 to 3, the detection performance significantly improves. That is because when $k=1$, too many clutters will be included in the spatio-temporal point cloud, which will cause the performance degradation of the proposed method. When $k$ increases from 3 to 15, the detection performance gradually decreases. That is because, with the value of $k$ increases, some dim targets will be missed, resulting in performance degradation. We choose $k=3$ as the threshold to balance performance and efficiency. 

\begin{table}
	% \vspace{3mm}
	\centering
	\renewcommand\arraystretch{1.1}
	\caption{The impact of different thresholds on detection performance. The best results are shown in \textbf{boldface}.}\label{thresholds}
	{
		\begin{tabular}{c|c|c|c|c|c}
			\hline
			 $k$ &Sampling ratio(\%) &Avg Re(\%) &Avg Pr(\%) &Avg F1(\%) &FPS\\[0.05cm]
			\hline
                1
                &10.56 &72.0	 &96.8	&82.3  &20.8
                \\
                3
                &1.22 &84.2	 &96.1	&89.7  & 98.8
                \\
			5   
                &0.27 &77.7	 &96.5	&85.8  &210.1
                \\
			  7        
                &0.12 &68.7	 &96.9	&79.8  &242.2
                \\
                9        
                &0.08 &62.6	 &97.1	&75.4  &246.8
                \\
                11        
                &0.06 &58.2	 &97.5	&72.1  &246.2
                \\
                13        
                &0.05 &54.0	 &97.6	&68.9  &247.0
                \\
                15        
                &0.04 &49.8	 &97.7	&65.3  &248.4
                \\
			\hline
	\end{tabular}}
 % \vspace{-2mm}
\end{table}

\textbf{(4) The Impact of Frame Numbers.} We analyze the performance of HiEUM with different frame numbers. Specifically, we evaluate the performance of HiEUM under different input frame numbers (i.e., 10, 20, 30, 40, 50, 60). The quantitative results are shown in Table \ref{frame_number}. It can be observed that when the frame number increases from 10 to 20, the detection performance is improved. That is because additional frames can provide long-term spatio-temporal information, which is beneficial for moving object detection. It is also notable that the detection performance tends to be degraded when the frame number is increased from 20 to 60 (the average F1 score drops from 89.7 to 88.8). That is because the sampling strategy will include more clutters into the point cloud due to the continuous movement of the satellite platform. Moreover, a further increase of frames can not provide performance improvement but brings extra computational burdens due to more clutters to be processed. Therefore, we utilize 20 frames as input to the proposed network.

\begin{table}
	% \vspace{-.05in}
	\centering
	\renewcommand\arraystretch{1.1}
	\caption{The impact of frame number on the detection performance. The best results are shown in \textbf{boldface}.}\label{frame_number}
	{
		\begin{tabular}{c|c|c|c|c}
			\hline
			Frames&Avg Re(\%) &Avg Pr(\%) &Avg F1(\%)  &FPS\\[0.05cm]
			\hline
			10    &72.4	&96.6	&82.6   &\textbf{108.7}\\
			20    &84.2	&96.1	&\textbf{89.7}  &98.8 \\
			30    &84.2	&95.2	&89.3   &96.8 \\
			40    &85.4	&94.5	&89.6  &95.1 \\
                50    &85.4	&94.0	&89.3  &91.2\\
			60    &84.8	&93.5	&88.8  &87.9 \\
			\hline
	\end{tabular}}
\end{table}

\textbf{(5) The Impact of Network Depth.} We analyze the performance of HiEUM with different network depths of the sparse U-net \cite{shi2020points}. Specifically, we evaluate the performance of HiEUM under different network depths (i.e., 2, 3, 4). The quantitative results are shown in Table \ref{layers}. It can be observed that when the layer number increases from 2 to 3, the detection performance is improved as the depth increases. That is because a network with more layers has a stronger modeling ability to achieve performance improvement. However, it is also notable that the detection performance tends to decrease when the layer number is increased from 3 to 4 (the average F1 score slightly drops). That is because more layers tend to be over-fitted on the limited training dataset. Therefore, we utilize a 3-layer sparse U-net as the backbone.

\begin{table}
	% \vspace{-.05in}
	\centering
	\renewcommand\arraystretch{1.1}
	\caption{The impact of network depth on the detection performance. The best results are shown in \textbf{boldface}.}\label{layers}
	{
		\begin{tabular}{c|c|c|c|c}
			\hline
			layers &Avg Re(\%) &Avg Pr(\%) &Avg F1(\%)  &FPS\\[0.05cm]
			\hline
			2    &83.8	&95.4	&89.2   &\textbf{132.11}\\
			3    &\textbf{84.2}	&\textbf{96.1}	&\textbf{89.7}  &98.83 \\
			4    &84.1	&96.0	&89.6   &92.58 \\
			\hline
	\end{tabular}}
 % \vspace{-2mm}
\end{table}

\section{Conclusion}
\label{conclusion}
In this paper, we present a generic highly efficient unsupervised framework for SVMOD. Under the proposed framework, we further proposed an effective and efficient sparse network to detect moving vehicles in spatio-temporal 3D point cloud representation, which can model long-term spatio-temporal information with much fewer computational burdens thanks to the intrinsic sparsity of moving vehicles. Extensive experiments have demonstrated the effectiveness and superior efficiency of our proposed method. 

\footnotesize
\bibliographystyle{IEEEtran}
\bibliography{reference.bib}

\end{document}